# Recognition of handwritten *Roman* Numerals using Tesseract open source OCR engine

**Sandip Rakshit [1], Amitava Kundu[2], Mrinmoy Maity[2], Subhajit Mandal[2], Satwika Sarkar[2], Subhadip Basu [2 #]**

[1] Techno India College of Technology, Kolkata, India
[2] Computer Science and Engineering Department, Jadavpur University, India

[#] Corresponding author. E-mail: subhadip@ieee.org

***Abstract***: *The objective of the paper is to recognize handwritten samples of Roman numerals using Tesseract open source Optical Character Recognition (OCR) engine. Tesseract is trained with data samples of different persons to generate one user-independent language model, representing the handwritten Roman digit-set. The system is trained with 1226 digit samples collected form the different users. The performance is tested on two different datasets, one consisting of samples collected from the known users (those who prepared the training data samples) and the other consisting of handwritten data samples of unknown users. The overall recognition accuracy is obtained as 92.1% and 86.59% on these test datasets respectively.*

## 1. INTRODUCTION

Any *Optical Character Recognition (OCR)* system eases the barrier of the keyboard interface between man & machine to a great extent, and help in office automation with huge saving of time and human effort. Such a system allows desired manipulation of the scanned text as the output is coded with ASCII or some other character code from the paper based input text. For a specific language based on some alphabet, OCR techniques are either aimed at printed text or handwritten text. The present work is aimed at the later.

Handwritten digit recognition is one of benchmark problem of research for the pattern recognition community. In general, OCR systems have potential applications in extracting data from filled in forms, interpreting handwritten addresses from postal documents for automatic routing, automatic reading of bank cheques etc. The key component of such application software is an OCR engine, enabled with the key functional modules like line extraction, line-to-word segmentation, word-to-character segmentation, character recognition and word-level lexicon analysis using standard dictionaries.

Development of an effective handwritten OCR engine with high recognition accuracy is a still an open problem for the research community. Lot of research efforts have already been reported [1-8] on different key aspects of handwritten character recognition systems. Tesseract is one such OCR engine covered under open source Apache licence 2.0, designed for recognition of printed characters of any script.In the current work, instead of developing a new handwritten OCR engine from scratch, we have used Tesseract 2.01 for recognition of handwritten numerals *Roman* script. Tesseract OCR engine provides high level of recognition accuracy on poorly printed or poorly copied dense text. In one of our earlier works [9], we had developed a system for estimation of recognition accuracy of Tesseract OCR engine on handwritten character samples of lower case *Roman* script collected from a single user. The performance of this OCR engine was also tested extensively on handwriting samples of multiple users in [10]. With these findings we can say that Tesseract open source Ocr Engine can be suitably tested and trained for recognition of handwritten characters as well.

In the current work, training samples are collected from three different users to generate a single user model for handwritten *Roman* numerals .The trained model is tested on fresh samples of the same users(who participated in the preparation of the training samples ) and also on handwritten samples of different other users. The objective of the work is to test the generality of the Tesseract OCR engine for recognition of digit patterns of handwritten *Roman* numerals using a user independent language model.

## 2. OVERVIEW OF THE TESSERACT OCR ENGINE

Tesseract is an open source (under Apache License 2.0) offline optical character recognition engine, originally developed at Hewlett Packard from 1984 to 1994. Tesseract was first started as a research project in HPLabs, Bristol [11-12]. In the year 1995 it is sent to UNLV where it proved its worth against the commercial engines of the time [13]. In the year 2005 Hewlett Packard and University of Nevada, Las Vegas, released it. Now it is partially funded by Google [14] and released under the Apache license, version 2.0. The latest version, Tesseract 2.03 is released in April, 2008. In the current work, we have used Tesseract version 2.01, released in August 2007.

Like any standard OCR engine, Tesseract is developed on top of the key functional modules like, line and word finder, word recognizer, static character classifier, linguistic analyzer and an adaptive classifier. However, it does not support document layout analysis, output formatting and graphical user interface. Currently, Tesseract can recognize printed text written in *English, Spanish, French, Italian, Dutch, German* and various other languages.

As for example to train Tesseract in *English* language 8 data files are required in tessdata sub directory. The 8 files used(also called a language set) for English are to be generated as follows:

    tessdata/eng.freq-dawg
    tessdata/eng.word-dawg
    tessdata/eng.user-words
    tessdata/eng.inttemp
    tessdata/eng.normproto
    tessdata/eng.pffmtable
    tessdata/eng.unicharset
    tessdata/eng.DangAmbigs

## 3. THE PRESENT WORK

In the current work, Tesseract 2.01 is used for recognition of handwriting samples of *Roman* numerals using an user independent language set. Key functional modules of the developed system are discussed the following sub-sections.

### 3.1. Collection of the dataset

For collection of the dataset for the current experiment, we have concentrated on hand written digits of *Roman* script only. Six handwritten document pages were collected from the three different users. For each user, two pages from each total six pages are collected for preparing the training set for the testing purposes, we have considered two types of users i.e known users and unknown users. Known users are the persons who participated in preparation of the training samples. For known user we consider two pages constituting the test dataset-1 (TD1)and for unknown user we consider three pages, i.e. the test dataset-2 (TD2).

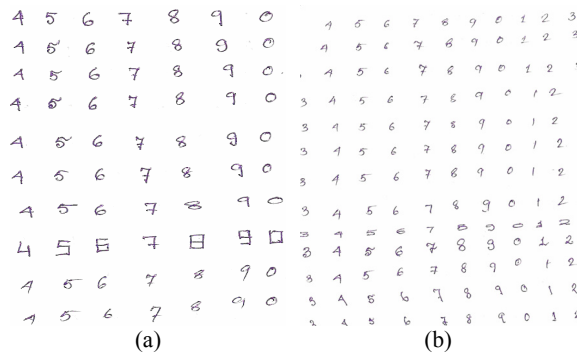

(a)                                   (b)

Fig. 1(a-b). Sample document pages containing training sets of isolated characters and free flow text

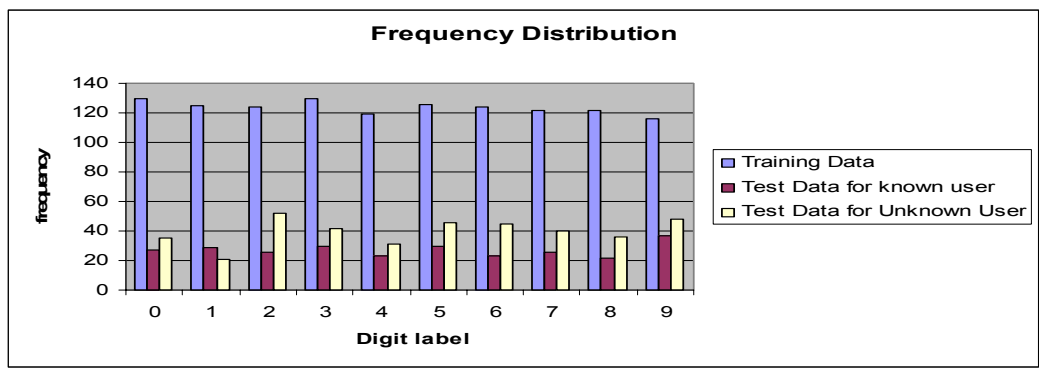

Fig. 2. Frequency distribution of different character samples during training

The training dataset contains around 1226 samples of isolated *Roman* digits collected from three different users .The test dataset 1 and test dataset 2 consists of 249 and 349 samples respectively collected from the known and the unknown users, the frequency of different digit samples in the training set and the two test sets is shown in Fig. 2.

**3.2. Labeling training data**

For labeling the training samples using Tesseract we have taken help of a tool named bbTesseract [14]. To generate the training files for a specific user, we need to prepare the box files for each training images using the following command:

*tesseract fontfile.tif fontfile batch.nochop makebox*

The box file is a text file that includes the characters in the training image, in order, one per line, with the coordinates of the bounding box around the image. The new Tesseract 2.01 has a mode in which it will output a text file of the required format. Some times the character set is different to its current training, it will naturally have the text incorrect. In that case we have to manually edit the file (using bbTesseract) to correct the incorrect characters in it. Then we have to rename fontfile.txt to fontfile.box. Fig. 3 shows a screenshot of the bbTesseract tool.

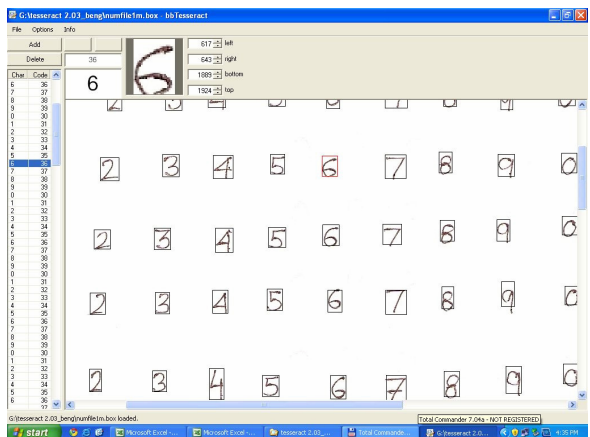

Fig. 3. A sample screenshot of a segmented training page using the bbTesseract tool

**3.3. Training the data using Tesseract OCR engine**

For training a new handwritten character set for any user, we have to put in the effort to get one good box file for a handwritten document page, run the rest of the training process, discussed below, to create a new language set. Then use Tesseract again using the newly created language set to label the rest of the box files corresponding to the remaining training images using the process discussed in section 3.2.

For each of our training image, boxfile pairs, run Tesseract in training mode using the following command:

*tesseract fontfile.tif junk nobatch box.train*

The output of this step is fontfile.tr which contains the features of each character of the training page. The character shape features can be clustered using the mftraining and cntraining programs:

*mftraining fontfile_1.tr fontfile_2.tr ...*

This will output three data files: inttemp , pffmtable and Microfeat, and the following command:

*cntraining fontfile_1.tr fontfile_2.tr ...*

This will output the normproto data file. Now, to generate the unicharset data file, unicharset_extractor program is used as follows:

*unicharset_extractor fontfile_1.box fontfile_2.box ...*

Tesseract uses 3 dictionary files for each language. Two of the files are coded as a Directed Acyclic Word Graph (DAWG), and the other is a plain UTF-8 text file. To make the DAWG dictionary files a wordlist is required for our language. The wordlist is formatted as a UTF-8 text file with one word per line. The corresponding command is:

*wordlist2dawg frequent_words_list freq-dawg*
*wordlist2dawg words_list word-dawg*

The third dictionary file name is user-words and is usually empty. The final data file of Tesseract is DangAmbigs file. This file cannot be used to translate characters from one set to another. The DangAmbigs file may be empty also.

Now we have to collect all the 8 files and rename them with a lang. prefix, where lang is the 3-letter code for our language and put them in our tessdata directory. Tesseract can then recognize text in our language using the command:

*tesseract image.tif output -l lang*

### 4. EXPERIMENTAL RESULTS

For conducting the current experiment, the training set is generated from isolated handwritten digit samples of three different users.As mentioned earlier , the test samples are collected in two datasets TD1 and TD2.All the samples are digitized using a flatbed scanner of resolution 300 dpi.

To estimate the performance of the present technique the following expression is developed.

$$\text{Recognition accuracy} = (C_t / (C_m + C_s))*100$$

where $C_t$ = the number of character segments producing true classification result and $C_m$ = the number of misclassified character segments and $C_s$ signifies the number of character Tesseract fails to segment, i.e., producing under segmentation. The rejected character/word samples are excluded from computation of recognition accuracy of the designed system.

To evaluate the generality of recognition capability of Tesseract OCR Engine over handwritten samples of different users, a single language set is prepared from all the training samples. The performance is tested on two different test datasets, viz., TD1 and TD2. Table-1 shows an analysis of recognition performance of the user independent dataset on the test data .As observed from the experiment, the overall recognition accuracy(excluding rejection samples) is obtained as 92.1% and 86.59% for TD1 and TD2 respectively. The rejection rates for the two datasets are observed as 10.4% and 10.8% respectively.

Table 1. Classification results on the test dataset

| Data set | Total characters | Misclass-ification | Rejec-tions | Success |
|---|---|---|---|---|
| TD1 | 249 | 20 | 26 | 92.1 |
| TD2 | 349 | 39 | 38 | 86.59 |

The reason behind high rejection rate of handwritten digit samples is because of the inability of Tesseract to segment digit samples into unique rectangular regions. Since these errors does not contribute to the recognition efficiency of the Tesseract OCR engine, we have excluded those samples from estimating the success rate of the user-independent language set for recognition of isolated handwritten *Roman* digit samples.

### 5. CONCLUSION

As observed from the experimental results, Tesseract OCR engine fares reasonably with respect to the core recognition accuracy on user-independent handwritten samples of *Roman* numerals. The performance of the system is validated on sample datasets of known and unknown users. A major drawback of the current technique is its failure to avoid segmentation errors in some of the digit samples. The performance of the designed system may be improved by incorporating more training samples, collected from users of varying age groups and professions. The system may further be extended for user-specific/user-independent recognition of handwritten characters of different Indic script.

In a nutshell, the developed technique shows enough promise in using an open source OCR engine for recognition of handwritten numerals of *Roman* script with a fairly acceptable accuracy.

## ACKNOWLEDGEMENTS

One of the authors, Mr. Sandip Rakshit is thankful to the authorities of Techno India College of Technology for necessary supports during the research work. Dr. Subhadip Basu is thankful to the "Center for Microprocessor Application for Training Education and Research", "Project on Storage Retrieval and Understanding of Video for Multimedia" of Computer Science & Engineering Department, Jadavpur University, for providing infrastructure facilities during progress of the work.